\documentclass{article}

\usepackage{arxiv}

\usepackage[utf8]{inputenc} % allow utf-8 input
\usepackage[T1]{fontenc}    % use 8-bit T1 fonts
\usepackage{hyperref}       % hyperlinks
\usepackage{url}            % simple URL typesetting
\usepackage{booktabs}       % professional-quality tables
\usepackage{amsfonts}       % blackboard math symbols
\usepackage{nicefrac}       % compact symbols for 1/2, etc.
\usepackage{microtype}      % microtypography
\usepackage{lipsum}		% Can be removed after putting your text content
\usepackage{graphicx}
\usepackage{natbib}
\usepackage{doi}

\usepackage{graphicx}
\usepackage{subcaption}
\usepackage{bm}
\usepackage{amssymb}
\usepackage{amsmath}
\usepackage{subcaption}
\usepackage{algorithm}
\usepackage{algorithmic}
\usepackage{mathtools}
\usepackage{xcolor}
\usepackage{wrapfig}
\usepackage{enumitem}
\usepackage{multirow}

\title{Labeled-Data-Free Meta-Learning: Efficient Task Generation Using Pre-trained Models and Unlabeled Data}

\author{
Lei Sun \\
Division of Information Science\\
Nara Institute of Science and Technology\\
Nara, Japan\\
\texttt{sun.lei.sn5@is.naist.jp}
\And
Yusuke Tanaka\\
NTT Communication Science Laboratories, NTT, Inc.\\
Kyoto, Japan\\
RIKEN Center for Advanced Intelligence Project\\
RIKEN, Saitama, Japan\\
\texttt{ysk.tanaka@ntt.com}
\And
Tomoharu Iwata\\
NTT Communication Science Laboratories, NTT, Inc.\\
Kyoto, Japan\\
\texttt{tomoharu.iwata@ntt.com}
}

\renewcommand{\shorttitle}{Labeled-Data-Free Meta-Learning}

\hypersetup{
pdftitle={Labeled-Data-Free Meta-Learning}
}

\begin{document}
\maketitle

\begin{abstract}
	Meta-learning without labeled data is crucial for real-world applications, where obtaining labeled datasets can be expensive or restricted due to privacy concerns. Data-Free Meta-Learning (DFML) addresses this challenge by leveraging pre-trained models without access to training data. However, existing DFML methods rely on model inversion to generate training data, a process that is generally difficult and computationally expensive due to the need of generating high-dimensional data matching the original distribution. To address this limitation, we propose a novel meta-learning setting that avoids model inversion by jointly leveraging pre-trained models and unlabeled data. Our method generates meta-training tasks by assigning soft labels from pre-trained models to unlabeled data. Since the quality of these tasks can vary, we introduce a task-weighting mechanism based on task confidence and class distribution balance to ensure effective meta-learning. Extensive experiments demonstrate that our approach substantially reduces computational cost and improves generalization, achieving up to 104$\times$ speedup and 8.4\%–36.4\% improvements in few-shot classification accuracy compared to state-of-the-art DFML methods.
\end{abstract}

\section{Introduction}
Meta-learning, also known as ``learning to learn'', enables models to leverage prior experience from solving diverse tasks, thereby facilitating the acquisition of inductive bias for the efficient learning of unseen but related tasks \citep{finn2017model, snell2017prototypical, li2017meta, iwata2020meta, sun2024meta, iwata2022sharing}. Traditional meta-learning approaches typically rely on a collection of tasks with labeled datasets. However, in many real-world scenarios, labeled data are often difficult or impossible to obtain due to concerns such as data privacy, security risks, and usage restrictions \citep{chen2019data, truong2021data}. Indeed, numerous individuals and institutions release task-specific pre-trained models on platforms like GitHub or Hugging Face, yet rarely provide access to their original training data.

To address this restriction, researchers have proposed Data-Free Meta-Learning (DFML) \citep{hu2023architecture, wei2024free, hu2023learning, wang2022meta}. DFML aims to extract knowledge directly from a collection of pre-trained models without requiring access to their original training data, enabling adaptation to unseen tasks. However, existing DFML approaches rely on model inversion to reconstruct training distributions, which requires high computational costs and results in degraded predictive accuracy. In particular, these methods typically train generators through model inversion \citep{frikha2023towards, patel2023learning} to synthesize images that approximate the original training distribution of each pre-trained model. Because this process involves generating high-dimensional data, it requires hundreds of iterative generate–forward–backward steps, resulting in an extremely time-consuming recovery process. Moreover, recovering the original training distribution through model inversion is generally difficult, as it cannot always generate high-quality data that matches the original distributions. Meta-learning with low-quality or mismatched data leads to error accumulation and degrades the performance of DFML methods.

\begin{figure}[t]
    \centering
    \includegraphics[width=0.6\linewidth]{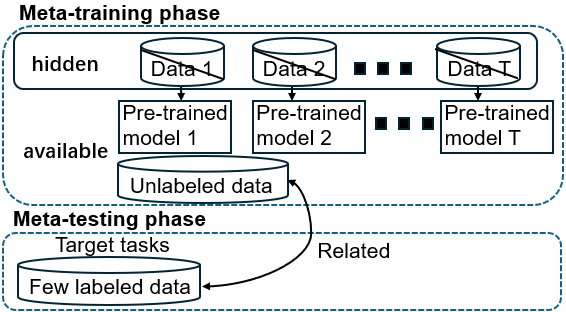}
    \caption{Illustration of the proposed setting. During the meta-training phase, only pre-trained models and unlabeled data related to the target tasks are available, while the original training datasets of the pre-trained models remain inaccessible. In the meta-testing phase, the model is adapted to target tasks using a few labeled instances.}
    \label{problem}
\end{figure}

To overcome this limitation, we propose a new problem setting that avoids model inversion by leveraging both pre-trained models and unlabeled data during the meta-training phase. Figure~\ref{problem} shows the proposed problem setting. Under this setting, we develop a meta-learning method that constructs meta-training tasks by assigning soft labels (i.e., probability distributions over classes), generated from multiple pre-trained models, to the unlabeled dataset. 

In the proposed setting, we assume that pre-trained models can be evaluated on the unlabeled instances to generate soft labels. We consider pre-trained models trained on classification tasks involving the same input modality and the same or related application domains as the target tasks, although their original training data may differ from the data used in the target tasks. The pre-trained models may also be heterogeneous in architecture, scale, and convergence levels.

Due to potential variations in the training domains and convergence levels of pre-trained models, the quality of the generated soft labels may differ. Consequently, directly utilizing tasks generated by these models may introduce noise into the meta-training process. Therefore, it is necessary to quantify the quality of each meta-training task. To address this issue, we propose a task-weighting strategy that combines two factors: (i) the average negative entropy of soft labels for all samples within a task, reflecting the overall task confidence; and (ii) the entropy of the mean soft label across samples in the task, indicating how evenly samples are distributed across classes. Specifically, higher values of factor (i) reflect greater task confidence, thereby providing clearer supervisory signals and enhancing stability during meta-training. However, excessively high negative entropy could indicate a concentration of samples within only one or a few classes, potentially causing class bias and negatively impacting model generalization. Therefore, we introduce factor (ii) to quantify class distribution balance, where higher values indicate a more uniform distribution across classes.

The proposed method avoids model inversion, thereby reducing computational costs, and introduces a task-weighting strategy that evaluates the usefulness of tasks generated by different pre-trained models, enabling more effective meta-training. Instead, the meta-training phase requires unlabeled data that are related, but not necessarily identical, to the target tasks. Here, “related” means that the unlabeled data come from the same input modality and application domain as the target tasks, although they may differ from the data used in the target tasks. This requirement is often practical because unlabeled data are generally easier to collect than labeled data. For example, in medical diagnosis, the target task may involve diagnosing specific diseases from only a few labeled examples, whereas hospitals can often collect large amounts of unlabeled medical images, such as X-rays or CT scans, through routine clinical practice. Although these unlabeled data may involve disease categories or patient populations different from those in the target tasks, they can still be considered related because they come from the same medical imaging domain.

The main contributions of this paper are as follows.
(i) We propose a meta-learning setting that enables the joint utilization of pre-trained models and unlabeled data, aligning more closely with real-world scenarios.
(ii) We propose a task construction method that avoids data recovery by leveraging soft labels generated by pre-trained models from an unlabeled dataset.
(iii) We propose a task-weighting mechanism that discerns the usefulness of tasks generated by various pre-trained models, resulting in effective meta-training algorithms.
(iv) Extensive experiments validate the superiority of our method across multiple benchmarks. 

\begin{table*}[t]
\centering
\caption{Comparison of the proposed setting with related problem settings. We compared the settings using four items. The first item (A) indicates whether learning can be performed without labeled training data during the corresponding training or adaptation phase. The second item (B) indicates whether unlabeled data can be used. The third item (C) indicates whether pre-trained models are used as available information sources. The last item (D) indicates whether the setting supports adaptation to unseen target tasks.}
\label{setting}
\begin{tabular}{lcccccc}
\hline
& DFML & UML & SFDA  & KD & KD-MAML &Ours \\
\hline
(A) Without labeled training data & \checkmark & \checkmark & \checkmark & \checkmark &&\checkmark \\
(B) Unlabeled data    &  & \checkmark & \checkmark & \checkmark &&\checkmark\\
(C) Pre-trained models &  \checkmark   &  & \checkmark & \checkmark &\checkmark &\checkmark\\
(D) Unseen target tasks & \checkmark & \checkmark &  &  & \checkmark &\checkmark\\
\hline
\end{tabular}
\end{table*}

\section{Related works}
Many meta-learning methods have been proposed \citep{finn2017model, snell2017prototypical, li2017meta, iwata2020meta, sun2024meta, iwata2022sharing}, but they typically rely on a collection of tasks with labeled datasets. In practice, labeled data are often costly or infeasible to obtain due to high costs and privacy concerns. Data-Free Meta-Learning (DFML) \citep{hu2023architecture, wei2024free, hu2023learning, wang2022meta} addresses this issue by leveraging multiple pre-trained models to enable adaptation to unseen tasks without accessing original training data. However, existing DFML approaches rely on model inversion-based data recovery, which is computationally expensive and fails to effectively utilize unlabeled data often available in real-world scenarios. 

Unsupervised Meta-Learning (UML) \citep{hsu2018unsupervised, khodadadeh2019unsupervised, jang2023unsupervised} constructs synthetic tasks from unlabeled data to acquire inductive bias for learning of unseen tasks.  These methods typically require a large-scale, unlabeled dataset, which is difficult to obtain in practice, especially in sensitive domains like healthcare and finance, where cross-institutional data sharing is heavily restricted. Moreover, UML approaches do not make use of existing pre-trained models. 

Source-Free Domain Adaptation (SFDA) \citep{mitsuzumi2024understanding, karim2023c, lee2022confidence} leverages a pre-trained model from the source domain and unlabeled data from the target domain to achieve domain adaptation. Although both SFDA and the proposed method leverage unlabeled data and pre-trained models to improve performance, existing SFDA methods are only applicable when the target task is predefined and cannot be directly extended to meta-learning frameworks.

Knowledge distillation (KD)~\citep{gou2021knowledge, wang2021knowledge, moslemi2024survey} trains a student model using the predictions of a pre-trained teacher model. Although both KD and our proposed method exploit soft labels generated by pre-trained models, standard KD generally aims to train the student to match the teacher’s predictions on a predefined task, rather than to enable adaptation to unseen tasks. Some studies have incorporated KD into meta-learning frameworks, such as KD-MAML~\citep{zhang2020knowledge}, which relies on labeled data to construct meta-training tasks. By contrast, our method constructs meta-training tasks from unlabeled data and multiple pre-trained models, without requiring access to the data originally used to train those models. Table~\ref{setting} summarizes the assumptions of each problem setting.

\begin{figure*}[t]
    \centering
    \includegraphics[width=\textwidth]{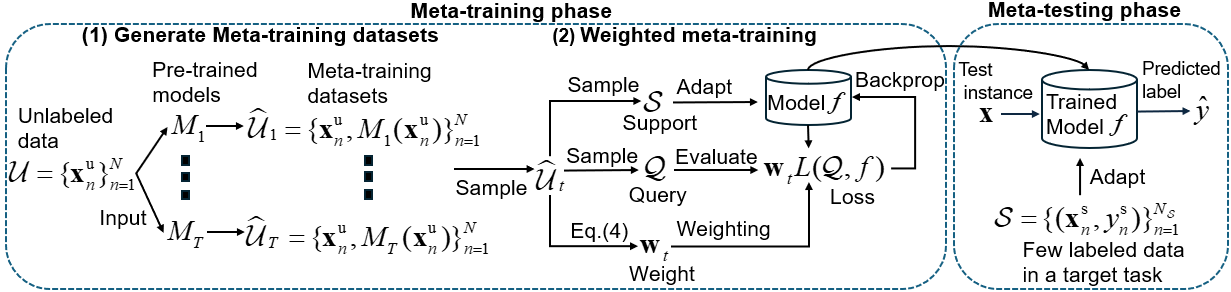}
    \caption{Illustration of the proposed method. At the meta-training phase, we generate meta-training datasets by assigning soft labels to the unlabeled dataset using pre-trained models. Task weights are then computed based on these soft labels, and the model is trained by minimizing the weighted meta-training loss. At the meta-testing phase, the trained model predicts the label of an input instance using a few labeled data.}
    \label{method}
\end{figure*}

\section{Problem formulation}\label{problem_formulation}
At the meta-training phase, we are given a collection of pre-trained models, denoted by $\mathcal{M} = \{M_{t}\}_{t=1}^{T}$, where each pre-trained model $M_{t}: \mathcal{X} \to \Delta^{C_{t}-1}$ performs classification over input space $\mathcal{X}$ by producing a probability distribution over $C_t$ classes. Here, $\Delta^{C_t - 1}$ denotes the set of non-negative $C_t$-dimensional vectors that sum to one. Note that the original training data for these models are not accessible and may differ from the data used in the target tasks. Additionally, we are given an unlabeled dataset $\mathcal{U} = \{\mathbf{x}_{n}^{\mathrm{u}}\}_{n=1}^{N}$, where each $\mathbf{x}_{n}^{\mathrm{u}}\in\mathcal{X}$ denotes the $n$-th input instance, and $N$ denotes the number of instances. This unlabeled dataset may differ from the data used in the target tasks, but it is assumed to come from the same input modality and application domain as the target tasks.

At the meta-testing phase, we are given a dataset on a target task, denoted as the support set $\mathcal{S} = \{(\mathbf{x}_{n}^{\mathrm{s}}, y_{n}^{\mathrm{s}})\}_{n=1}^{N_{\mathcal{S}}}$, where each $\mathbf{x}_{n}^{\mathrm{s}}\in\mathcal{X}$ denotes the $n$-th input instance, $y_{n}^{\mathrm{s}}$ represents the label corresponding to $\mathbf{x}_{n}^{\mathrm{s}}$. Here, the number of instances $N_{\mathcal{S}}$ is small. Our aim is to improve the test label prediction performance in target tasks.  

Overall, the original training data of the pre-trained models, the unlabeled data, and the data used in the target tasks are assumed to share the same input modality and to be from the same or related application domains, although they may differ from one another.

\section{Proposed method}\label{method_proposed}
We propose a meta-learning method that learns from both pre-trained models and unlabeled data. Our method aims to construct a set of meta-training tasks by utilizing unlabeled data and pre-trained models. To achieve this, we generate meta-training datasets by assigning soft labels to unlabeled dataset $\mathcal{U}$ using pre-trained models $\mathcal{M}$. The datasets for meta-training tasks are denoted by
$\hat{\mathcal{U}} = \{ \hat{\mathcal{U}}_t \}_{t=1}^{T}$,
where
\begin{equation}\label{y}
\hat{\mathcal{U}}_t  = \left\{ (\mathbf{x}_{n}^{\mathrm{u}}, M_{t}(\mathbf{x}_{n}^{\mathrm{u}})) \right\}_{n=1}^{N},
\end{equation}
is the $t$-th meta-training dataset, and $M_t(\mathbf{x}_{n}^{\mathrm{u}})$ denotes the soft label generated by the $t$-th pre-trained model for input $\mathbf{x}_{n}^{\mathrm{u}}$ and is defined as
\begin{equation}
M_t(\mathbf{x}_{n}^{\mathrm{u}}) = \left[ M_t^{1}(\mathbf{x}_{n}^{\mathrm{u}}), M_t^{2}(\mathbf{x}_{n}^{\mathrm{u}}), \cdots, M_t^{C_{t}}(\mathbf{x}_{n}^{\mathrm{u}})\right].
\end{equation}
Here, $M_t^{c}(\mathbf{x}_{n}^{\mathrm{u}})$ denotes the predicted probability that input $\mathbf{x}_{n}^{\mathrm{u}}$ belongs to class $c$ according to the $t$-th pre-trained model, with $\sum_{c \in C_{t}} M_t^{c}(\mathbf{x}_{n}^{\mathrm{u}}) = 1$ and $M_t^{c}(\mathbf{x}_{n}^{\mathrm{u}}) \geq 0$ for all $c$.

Tasks generated by multiple pre-trained models vary in usefulness for meta-training. To quantify this, we introduce the following task weights based on the soft labels:
{
\begin{equation}\label{weight}
{w}_{t}
= \exp\left(-\alpha\frac{1}{N_{}}\sum\limits_{n=1}^{N_{}}H(M_{t}(\mathbf{x}_n^{\mathrm{u}})) + \beta H (\frac{1}{N_{}}\sum\limits_{n=1}^{N_{}}M_{t}(\mathbf{x}_n^{\mathrm{u}}))\right),
\end{equation}}%
where $H(\cdot)$ is the entropy, $\alpha$ and $\beta$ are hyperparameters that balance the contributions of two entropy-based components in the weight computation, and $\alpha, \beta \in \mathbb{R}_{\geq 0}$. Our task-weighting strategy computes each task weight by combining two factors: (i) the negative entropy of soft labels averaged over all samples, which measures predictive uncertainty; and (ii) the entropy of the mean soft label across all samples in the task, which measures how uniformly the unlabeled samples are distributed across classes. Specifically, for (i), a higher average negative entropy indicates lower predictive uncertainty; thus, pre-trained models that are more relevant to the target task are expected to produce lower uncertainty predictions on unlabeled data. For (ii), the entropy of the mean soft label quantifies how uniformly samples are distributed across classes. Under the assumption that the target task maintains a balanced class distribution, this term reflects class-distribution alignment. Based on these two factors, we assign higher weights to tasks exhibiting both lower predictive uncertainty and a more uniform predicted class distribution, as these tasks are more likely to align with the target task.

%\begin{equation}\label{obj_func}
%\min_{\theta} \mathbb{E}_{\hat{\mathcal{U}}_{t}\sim\hat{\mathcal{U}}}\left[{w}_{t} \mathbb{E}_{(\mathcal{S},\mathcal{Q})\sim \hat{\mathcal{U}}_{t}}L^{outer} \left(\mathcal{Q}_{t}, \theta_{t}\right)\right], \quad s.t. \quad \theta_{t} = \theta - \gamma\nabla_{\theta} L^{inner} (\mathcal{S}_{t}, \theta)
%\end{equation}

%\begin{equation}\label{obj_func}
%\min_{\theta} \mathbb{E}_{\hat{\mathcal{U}}_{t}\sim\hat{\mathcal{U}}}\left[{w}_{t} \mathbb{E}_{(\mathcal{S},\mathcal{Q})\sim \hat{\mathcal{U}}_{t}}L(\mathcal{Q}_{t}|\mathcal{S}_{t};\theta)\right],
%\end{equation}

We consider model $f(\cdot;\psi,\theta)$ with task-shared parameters $\theta$ and task-specific parameters $\psi$, which is trained by a bi-level optimization framework. Task-shared parameters $\theta$ are optimized during the outer optimization, task-specific parameters $\psi$ are adapted through the inner optimization. Model $f(\mathbf{x};\psi,\theta)$ outputs the $C_{t}$-dimensional probability distribution over classes for input $\mathbf{x}$. Once the meta-training tasks and their corresponding weights are determined, the objective to be minimized is 
the following expected weighted test error:
\begin{equation}\label{obj_func}
\min_{\theta} \mathbb{E}_{\hat{\mathcal{U}}_{t}\sim\hat{\mathcal{U}}}\left[{w}_{t} \mathbb{E}_{(\mathcal{S},\mathcal{Q})\sim \hat{\mathcal{U}}_{t}}L^{\mathrm{outer}}(\mathcal{Q}, f(\cdot,\psi_{\mathcal{S}},\theta))\right], \quad {\rm s.t.} \quad \psi_{\mathcal{S}} = \arg\min_{\psi} L^{\mathrm{inner}}(\mathcal{S}, f(\cdot,\psi,\theta)),
\end{equation}
where $\mathbb{E}$ denotes the expectation. $L^{\mathrm{inner}}(\mathcal{S}, f(\cdot,\psi,\theta))$ is the inner loss used to adapt $\psi$ on support set $\mathcal{S}$, while $L^{\mathrm{outer}}(\mathcal{Q}, f(\cdot,\psi,\theta))$ is the outer loss used to evaluate the generalization performance of $\theta$ on query set $\mathcal{Q}$. The support and query sets are subsets of
$\hat{\mathcal{U}}_{t}$ and are disjoint, i.e.,
$\mathcal{S}\cap\mathcal{Q}=\emptyset$. Each set contains pairs of input instances and soft labels, which are randomly sampled from $\hat{\mathcal{U}}_{t}$. When the meta-training tasks constructed from unlabeled data and pre-trained models are related to the target tasks, training the task-shared parameters using bi-level optimization on these tasks can enhance the model's performance on the target tasks. However, due to variations among pre-trained models, the quality of soft labels in the generated meta-learning tasks may be inconsistent. To address this, our proposed task-weighting mechanism can assign weights to each task based on task confidence and class balance, enabling the model to discern the relative usefulness of different tasks. By minimizing the weighted meta-test error defined in Eq.~\eqref{obj_func}, our approach achieves more effective meta-training.

In outer optimization, we use the following loss:
\begin{equation}\label{query_loss}
L^{\mathrm{outer}}(\mathcal{Q}, f(\cdot;\psi_{\mathcal{S}},\theta))  
=\frac{1}{N_{\mathrm{Q}}}\sum\limits_{(\mathbf{x},M_{t}(\mathbf{x}))\in \mathcal{Q}}\mathsf{KL}\left(M_{t}(\mathbf{x}), f(\mathbf{x}; \psi_{\mathcal{S}},\theta)\right),
\end{equation}
where $\mathsf{KL}(\cdot)$ is the Kullback-Leibler divergence. By formulating the outer objective as the discrepancy between the predictions generated by the pre-trained models and those produced by model $f$, task-shared knowledge can be effectively transferred from pre-trained models to model $f$.

In inner optimization, we find task-specific parameters adapted to the support set with soft labels. As soft labels carry more informative content than hard labels, their use in inner optimization enables the learning of more effective task-specific parameters. The overall meta-training procedure of our model is summarized in Algorithm~\ref{alg}. In practice, the model parameters can be trained using various meta-learning algorithms \citep{vinyals2016matching, bertinetto2018meta}. In the following paragraph, we use Prototypical Networks \citep{snell2017prototypical} as an illustrative example of the optimization process. In the meta-testing phase, the meta-learned model $f$ is adapted to the target task through inner optimization with the support set $\mathcal{S}$, after which the adapted model is employed to predict the labels of the query instances. Figure~\ref{method} illustrates the proposed method.

\begin{algorithm}[t]
\caption{Meta-learning procedures.}\label{alg}
\textbf{Input}: Pre-trained models $\mathcal{M} = \{M_{t}\}_{t=1}^{T}$, unlabeled dataset $\mathcal{U}=\{\mathbf{x}_{n}\}_{n=1}^{N}$,
number of support instances $N_{\mathrm{S}}$, number of query instances $N_{\mathrm{Q}}$, batch size $B$.

\textbf{Output}: Trained model parameters $\theta$
\begin{algorithmic}[1] %[1] enables line numbers
\STATE Randomly initialize $\theta$
\STATE Create meta-training datasets $\hat{\mathcal{U}}= \{\hat{\mathcal{U}}_{t}\}_{t=1}^{T}$ by Eq.~\eqref{y} using $\mathcal{U}$ and $\mathcal{M}$. 
\STATE Compute weight $\{{w}_{t}\}_{t=1}^{T}$ by Eq.~\eqref{weight} using $\hat{\mathcal{U}}$
\WHILE{not done}
\STATE Initialize loss $J \gets 0$
\STATE Randomly sample a batch of tasks $\mathcal{B}$ from $\{1,\ldots,T\}$

\FOR{ $t \in \mathcal{B}$}

\STATE Randomly sample support set $\mathcal{S}$ with size $N_{\mathrm{S}}$ from $\hat{\mathcal{U}}_{t}$

\STATE Randomly sample query set $\mathcal{Q}$ with size $N_{\mathcal{Q}}$ from $\hat{\mathcal{U}}_{t}\setminus \mathcal{S}$

\STATE Obtain task-specific parameter $\psi_{\mathcal{S}}$ via inner optimization: $\psi_{\mathcal{S}}=\arg\min\limits_{\psi}L^{\mathrm{inner}}(\mathcal{S}, f(\cdot;\psi,\theta))$

\STATE Calculate weighted outer loss by Eq.~\eqref{query_loss} and ${w}_{t}$ , $J\gets J+{w}_{t}L^{\mathrm{outer}}(\mathcal{Q}, f(\cdot;\psi_{\mathcal{S}},\theta))$
\ENDFOR
\STATE  Update model parameters $\theta$ using gradient of loss $J/B$
\ENDWHILE
\end{algorithmic}
\end{algorithm}

\paragraph{Illustrative example based on Prototypical Networks.}
In Prototypical Networks, the outer optimization aims to learn an embedding function $\phi(\cdot; \theta)$ with task-shared parameters $\theta$, and task-specific parameters $\psi = \{\psi_{c}\}_{c=1}^{C_{t}}$ are a set of prototype for each class. Specifically, model $f$ is defined by
\begin{equation}\label{predict}
f^{c}(\mathbf{x};\psi_{\mathcal{S}},\theta)=
\frac{\exp\left(-\frac{1}{2}||(\phi(\mathbf{x};\theta) - \mathbf{\psi}_{\mathcal{S}}^{c})||^{2}\right)}{\sum\limits_{c'}\exp\left(-\frac{1}{2}||(\phi(\mathbf{x};\theta)- \mathbf{\psi}_{\mathcal{S}}^{c'})||^{2}\right)}.
\end{equation}

Each prototype $\psi_{\mathcal{S}}^{c}$ adapted to the support set
is computed as the weighted mean of the embedded support instances, where the weights correspond to the soft label probabilities of those instances belonging to class $c$:
\begin{equation}
\psi_{\mathcal{S}}^{c}
= \arg\min_{\psi^{c}} \left[
-\sum_{\mathbf{x} \in \mathcal{S}} M_{t}^{c}(\mathbf{x}) 
   \log \mathcal{N}(\phi(\mathbf{x};\theta);\psi^c,I)\right] 
= \frac{\sum\limits_{\mathbf{x}\in \mathcal{S}}M_{t}^{c}(\mathbf{x}) \phi(\mathbf{x};\theta)}
        {\sum\limits_{\mathbf{x}\in \mathcal{S}}M_{t}^{c}(\mathbf{x})},
\end{equation}
where the weighted negative log likelihood with Gaussian is used as the inner loss,
and $\mathcal{N}(\cdot; \psi^{c}, I)$ is the probability density function of a multivariate Gaussian distribution with mean vector $\psi^{c}$ and identity covariance matrix $I$.

\section{Experiments}

\subsection{Datasets and pre-trained models}\label{setup}
We conducted experiments on three widely-used few-shot learning benchmark datasets: Omniglot \citep{lake2011one}, miniImageNet \citep{vinyals2016matching}, and tieredImageNet \citep{ren2018meta}. Following standard splits \citep{wertheimer2021few}, we split each dataset into the meta-training, meta-validation, and meta-testing subsets with disjoint label spaces. Here, we further divided the meta-testing dataset into two subsets with disjoint label spaces: one for sampling unlabeled data used during meta-training, and the other for evaluation during meta-testing. This ensured that the unlabeled data were drawn from tasks that are distinct from those used during meta-testing. 

For the pre-trained models, we followed the DFML assumption that the meta-training data are inaccessible. For miniImageNet and tieredImageNet, following prior work \citep{hu2023architecture, wei2024free, hu2023learning, wang2022meta}, we collected 100 pre-trained models, each trained on an $N$-way classification task sampled from the meta-training subset. For Omniglot, we first meta-trained the models using MAML \citep{finn2017model} on the meta-training subset and subsequently fine-tuned them on N-way classification tasks sampled from the same subset. This results in 100 fine-tuned pre-trained models. Through this procedure, we collected a pool of pre-trained models covering a broad range of accuracies on each dataset. The miniImageNet pool had an average accuracy of 78.3\%, with a minimum of 31.0\% and a maximum of 92.4\%, and included 3 models with accuracies below 50\%.
The tieredImageNet pool averaged 72.5\% accuracy, ranging from 29.2\% to 89.6\%, and contained 7 models below 50\%. For Omniglot (5-way), the pool had an average accuracy of 90.7\%, with accuracies between 59.7\% and 99.1\%, and 0 models fell below 50\%. For Omniglot (20-way), the pool averaged 75.3\%, with a minimum of 47.2\% and a maximum of 94.2\%, and included one model with accuracies below 50\%.

In the multi-domain scenario, we evaluated the proposed method on the CUB \citep{WahCUB_200_2011} and CIFAR-FS \citep{bertinetto2018meta} datasets. The data splits were kept consistent with the standard scenario. The key difference was that, in the multi-domain setting, we collected a total of 100 pre-trained models, with 50 trained on miniImageNet and the other 50 trained on CUB or CIFAR-FS.

\subsection{Implementation details}
For the model architecture, we adopted Conv4 \citep{finn2017model, snell2017prototypical} as the architecture of the meta-learner and the pre-trained models for a fair comparison with existing works. For hyperparameters, batch size $B$ was set to 100, Adam \citep{kingma2014adam} was used as the optimizer, with the learning rate set to 0.001, and balance factors $\alpha$ and $\beta$ were set to 0.5 and 0.5. The experimental results under varying hyperparameters are shown in the following sections. Number of instances $N$ in unlabeled dataset $\mathcal{U}$ was set to 100 for the 5-way setting and 400 for the 20-way setting. In each meta-training task constructed according to Eq.~\eqref{y}, the support set contained 25 samples, and the query set contained 75 samples for the 5-way setting, and 100 and 300 samples for the 20-way setting, respectively. The samples for the support and query sets were randomly selected from the dataset constructed according to Eq.~\eqref{y}. Results are averaged over five independent runs with different unlabeled datasets and model initializations. Each run is evaluated on 600 meta-test tasks. We report the mean ± standard error. All methods are evaluated under the same meta-test protocol and using the same support/query splits. All training times were measured using an NVIDIA TITAN RTX GPU. 

\subsection{Comparative methods}
We compared the proposed method with the following baselines and competing methods.
\textbf{(i) Random:} model parameters were randomly initialized and optimized from scratch using only the support set of each meta-test task. \textbf{(ii) K-nearest neighbors (embedded):} samples were embedded using the pre-trained model with the highest validation accuracy on the original training task, and query labels were predicted by nearest-neighbor classification based on Euclidean distances in the embedding space. \textbf{(iii) K-nearest neighbors (pixel):} query labels were predicted by nearest-neighbor classification using Euclidean distances computed directly in the raw pixel space without embedding. \textbf{(iv) Best pre-trained model:} the pre-trained model was selected according to its validation accuracy on the original training task, fine-tuned on the support set of each meta-test task, and then used to predict the query labels. \textbf{(v) Pseudo-supervised contrastive learning (PsCo) \citep{jang2023unsupervised}:}  a state-of-the-art unsupervised meta-learning method based on contrastive learning with a momentum network. 
\textbf{(vi) Fast and better data-free meta-learning (FREE) \citep{wei2024free}:} a state-of-the-art data-free meta-learning method that recovers meta-training tasks from pre-trained models and employs gradient alignment for meta-training.
\textbf{(vii) Knowledge distillation for model-agnostic meta-learning (KD-MAML) \citep{zhang2020knowledge}:} A meta-learning method based on knowledge distillation. Since the original method requires labeled data during the meta-training phase, we adapted it to our setting by training it on the meta-training tasks constructed via Eq.~\eqref{y}. Under this task construction, KD-MAML was treated as optimizing Eq.~\eqref{obj_func} without task weighting, while the inner-loop optimization was performed using hard pseudo-labels.

Methods~(i)--(iv) did not involve meta-training and used only the support set of each meta-test task for task-specific adaptation or prediction. Method~(v) used only the unlabeled dataset for meta-training, while Method~(vi) used only the pre-trained models. Method~(vii) was trained using both the unlabeled dataset and the pre-trained models. The proposed method~(Ours) was trained under the same setting as in Method~(vii).

\begin{table*}[t]
  \centering
  \caption{Average accuracies (\%) on the Omniglot dataset. Results are averaged over five independent runs with different unlabeled datasets, with standard errors reported. Methods without unlabeled data report only average accuracies. \textbf{Bold} indicates the best and statistically comparable results according to the paired $t$-test ($p=0.05$).}
  \label{omniglot}
  \begin{tabular}{lcccc}
\hline
    & \multicolumn{4}{c}{Omniglot (way, shot)}  \\
    Method & (5,1) & (5,5) & (20,1) & (20,5)  \\
\hline
    Random & 50.2 & 73.6 & 23.9 & 43.8  \\
    K-nearest neighbors (embedded)  & 41.5 & 52.3 & 36.4 & 52.8\\
    K-nearest neighbors (pixel)  & 44.1 & 67.1 & 25.2    & 47.1\\
    Best pre-trained model  & 59.8 & 79.5 & 55.3 & 73.1 \\
\hline
    PsCo    & 54.9 $\pm$ 0.2 & 65.9 $\pm$ 0.2 & 31.2  $\pm$ 0.5 & 43.4 $\pm$ 0.7 \\
\hline
    FREE              & 41.3 & 55.4 & 32.1 & 48.9  \\
\hline
    KD-MAML & 58.8 $\pm$ 0.6 & 72.4 $\pm$ 2.7 & 55.2 $\pm$ 0.9 & 66.1 $\pm$ 1.3 \\
    Ours  & \textbf{72.5 $\pm$ 0.7} & \textbf{87.5 $\pm$ 1.8} & \textbf{65.1 $\pm$ 1.3} & \textbf{85.3 $\pm$ 0.7}\\
\hline
  \end{tabular}
\end{table*}

\begin{table*}[t]
  \centering
  \caption{
  Average accuracies (\%) on the miniImageNet and tieredImageNet dataset. Results are averaged over five independent runs with different unlabeled datasets, with standard errors reported. Methods without unlabeled data report only average accuracies. \textbf{Bold} indicates the best and statistically comparable results according to the paired $t$-test ($p=0.05$).}
\label{imagenet}
  \begin{tabular}{lcccc}
\hline
    & \multicolumn{2}{c}{miniImageNet (way, shot)}& \multicolumn{2}{c}{tieredImageNet (way, shot)} \\
    Method & (5,1) & (5,5) & (5,1) & (5,5)  \\
    %\midrule
    %\rowcolor{gray!15}
    %\multicolumn{5}{c}{\text{Fine-tune on the meta-testing support set without meta-training}} \\
\hline
    Random & 28.1 & 37.2 & 29.6 & 38.2  \\
    K-nearest neighbors (embedded)  & 28.8 & 39.2 & 31.7 & 32.7\\
    K-nearest neighbors (pixel)  & 27.7 & 30.4 & 25.7 & 29.2\\
    Best pre-trained model  & 24.5& 31.6 & 23.7 & 28.5 \\
    %\midrule
    %\rowcolor{gray!15}
    %\multicolumn{5}{c}{\text{Meta-train using only unlabeled data}} \\
\hline
    PsCo            & 31.7 $\pm$ 0.4 & 38.5 $\pm$ 1.8 & 30.2 $\pm$ 0.2 & 30.5 $\pm$ 0.4  \\
    %\midrule
    %\rowcolor{gray!15}
    %\multicolumn{5}{c}{\text{Meta-train using only pre-trained models}} \\
\hline
    FREE              & 25.2 & 35.8 & 23.7 & 35.1  \\
    %\midrule
    %\rowcolor{gray!15}
    %\multicolumn{5}{c}{\text{Meta-train using both unlabeled data and pre-trained models}} \\
\hline
    KD-MAML & 28.3 $\pm$ 1.7 & 36.3 $\pm$ 2.2 & 28.5  $\pm$ 1.9& 33.1  $\pm$ 1.2\\
    %Ours (without weight) & 34.2 $\pm$ 2.2 & 46.3 $\pm$ 1.6 & 32.9 $\pm$ 0.5 & 46.0 $\pm$ 1.0 \\
    Ours  & \textbf{33.6 $\pm$ 1.2 } & \textbf{48.7 $\pm$ 0.9} & \textbf{34.8 $\pm$ 0.7} & \textbf{48.1 $\pm$ 0.6}\\
\hline
  \end{tabular}
\end{table*}

\begin{table}[t]
\caption{Training computational time in hours on miniImageNet under the 5-way 1-shot setting.}
\label{training_time}
\centering
\begin{tabular}{cccc}
\hline
FREE & PsCo & KD-MAML & Ours \\
10.4 $\pm$ 0.1 & 1.5 $\pm$ 0.1 &0.3 $\pm$ 0.0 & 0.1 $\pm$ 0.0\\
\hline
\end{tabular}
\end{table}

\begin{table*}[t]
  \centering
  \caption{
  Average accuracies (\%) on CUB and CIFAR-FS in the multi-domain scenario. Results are averaged over five independent runs with different unlabeled datasets, with standard errors reported. Methods without unlabeled data report only average accuracies. \textbf{Bold} indicates the best and statistically comparable results according to the paired $t$-test ($p=0.05$).}
\label{cross_domain}
  \begin{tabular}{lcccc}
\hline
    & \multicolumn{2}{c}{CUB (way, shot)}& \multicolumn{2}{c}{CIFAR-FS (way, shot)} \\
    Method & (5,1) & (5,5) & (5,1) & (5,5)  \\
    %\midrule
    %\rowcolor{gray!15}
    %\multicolumn{5}{c}{\text{Meta-train using only unlabeled data}} \\
\hline
    PsCo   & 31.8 $\pm$ 1.7 & 36.1 $\pm$ 0.9 & 32.5 $\pm$ 0.5 & 39.8  $\pm$ 1.1\\
    %\midrule
    %\rowcolor{gray!15}
    %\multicolumn{5}{c}{\text{Meta-train using only pre-trained models}} \\
\hline
    FREE & 24.5  &34.0  &26.2  &37.6   \\
    %\midrule
    %\rowcolor{gray!15}
    %\multicolumn{5}{c}{\text{Meta-train using both unlabeled data and pre-trained models}} \\
\hline
    KD-MAML &  30.0 $\pm$ 1.5 &  35.1 $\pm$ 1.2  & 31.6  $\pm$ 2.3 & 40.5 $\pm$ 2.5 \\
    %Ours (without weight) & 30.3 $\pm$ 2.5 & 37.3 $\pm$ 1.3 & 33.1 $\pm$ 1.1 & 44.5 $\pm$ 0.9  \\
    Ours  & \textbf{34.7 $\pm$ 1.3} & \textbf{44.8 $\pm$ 1.4 } & \textbf{35.3 $\pm$ 1.2 } & \textbf{45.1 $\pm$ 0.1}\\
\hline
  \end{tabular}
\end{table*}

\begin{table}[t]
  \centering
  \caption{
  Average accuracies (\%) on miniImageNet under cross-architecture scenario. Results are averaged over five independent runs with different unlabeled datasets, with standard errors reported. Methods without unlabeled data report only average accuracies. \textbf{Bold} indicates the best and statistically comparable results according to the paired $t$-test ($p=0.05$).}
  \label{cross_archi}
  \begin{tabular}{lccccccc}
    \hline
    %& \multicolumn{2}{c}{miniImageNet (way, shot)} \\
    %\cmidrule(lr){2-3}
    Method & (5-way, 1-shot) &(5-way, 5-shot)  \\
    \hline
    FREE   & 24.1 & 33.9  \\
    KD-MAML   & 27.0 $\pm$ 0.9 & 32.6 $\pm$ 1.2 \\
    Ours   & \textbf{33.1 $\pm$ 2.3} & \textbf{48.1$\pm$ 2.0}  \\
    \hline
  \end{tabular}
\end{table}

\begin{table*}[t]
\caption{Ablation study. Average accuracies and standard errors (\%) under miniImageNet 5-way 1-shot setting. We compare the proposed method (ours) with the following variants: removing factor (i) in task-weighting ($\alpha=0, \beta=1$), removing factor (ii) ($\alpha=1, \beta=0$), removing both factors (w/o weight) and replace soft labels with hard labels in the support (inner hard), query (outer hard).}
\centering
\label{ablation}
\begin{tabular}{cccccc}
\hline
w/o weight & $\alpha=0, \beta =1$& $\alpha=1, \beta =0$&inner hard & outer hard   & Ours\\
\hline
31.0 $\pm$ 2.9 & 31.9 $\pm$ 3.1 & 32.7 $\pm$ 2.5 &  31.1 $\pm$ 3.3 & 29.1 $\pm$ 2.4 &\textbf{33.6 $\pm$ 1.2 }\\
%miniImageNet (5,5) & 46.3  $\pm$ 1.6   &  46.8 $\pm$ 4.1 & 45.3 $\pm$ 3.8 & 43.9  $\pm$ 3.6 & \textbf{50.4 $\pm$ 1.4}\\
\hline
\end{tabular}
\end{table*}

\begin{figure*}[t]
    \centering

    \begin{subfigure}[t]{0.32\textwidth}
        \centering
        \includegraphics[width=\linewidth]{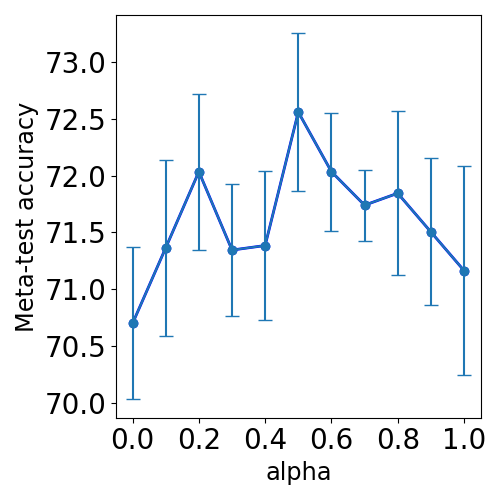}
        \caption{Omniglot 5-way}
        \label{fig:omni5_alpha}
    \end{subfigure}
    \begin{subfigure}[t]{0.32\textwidth}
        \centering
        \includegraphics[width=\linewidth]{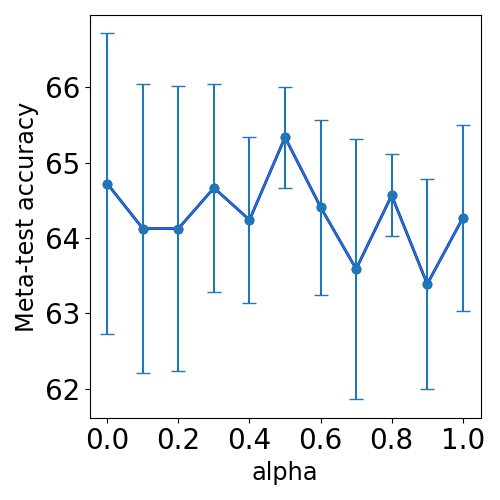}
        \caption{Omniglot 20-way}
        \label{fig:omni20_alpha}
    \end{subfigure}

    \vspace{0.8em}
    
    \begin{subfigure}[t]{0.32\textwidth}
        \centering
        \includegraphics[width=\linewidth]{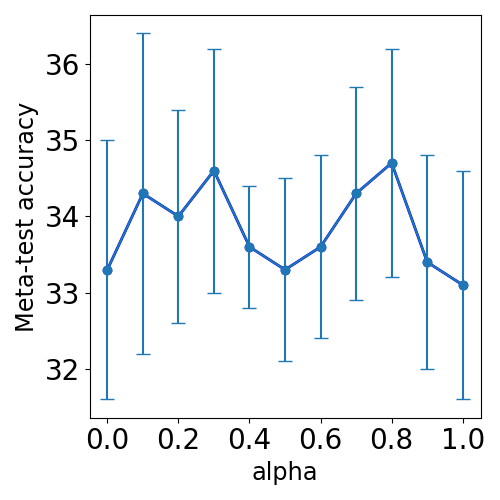}
        \caption{miniImageNet}
        \label{fig:mini_alpha}
    \end{subfigure}
    \begin{subfigure}[t]{0.32\textwidth}
        \centering
        \includegraphics[width=\linewidth]{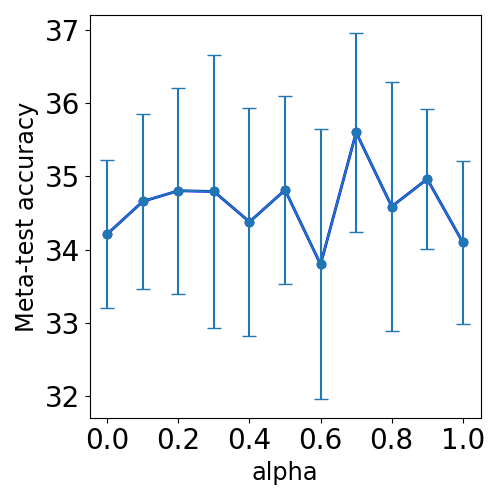}
        \caption{tieredImageNet}
        \label{fig:tiered_alpha}
    \end{subfigure}

    \caption{
    Analysis of the balance factors $\alpha$ and $\beta$ in the proposed task-weighting mechanism under the 1-shot setting.
    Meta-testing accuracies are reported by varying $\alpha$ with $\beta = 1-\alpha$ on (a) Omniglot 5-way, (b) Omniglot 20-way, (c) miniImageNet 5-way, and (d) tieredImageNet 5-way tasks.
    Bars indicate the standard error.
    }
    \label{fig:alpha_beta}
\end{figure*}

\begin{figure*}[t]
    \centering
    
    \begin{subfigure}[t]{0.32\textwidth}
        \centering
        \includegraphics[width=\linewidth]{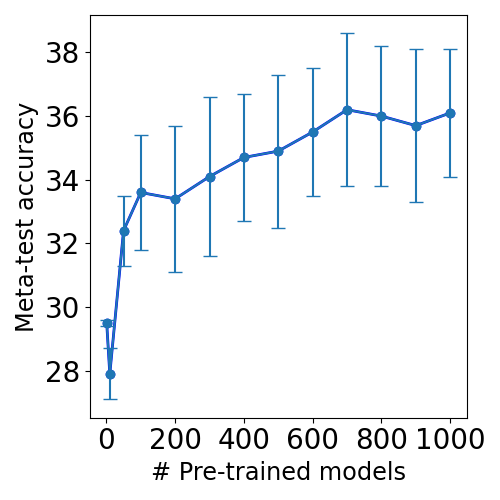}
        \caption{Number of pre-trained models}
        \label{fig:num_models}
    \end{subfigure}
    \begin{subfigure}[t]{0.32\textwidth}
        \centering
        \includegraphics[width=\linewidth]{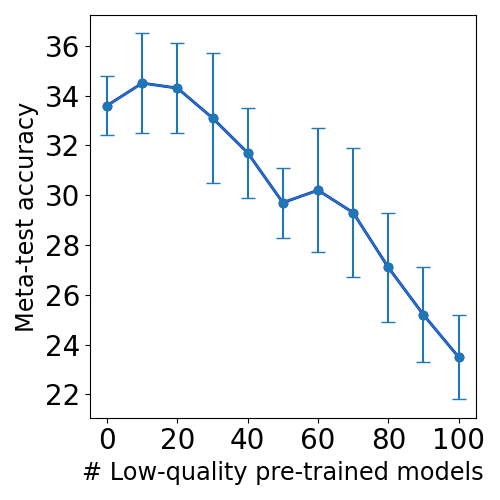}
        \caption{Number of low-quality models}
        \label{fig:noisy_models}
    \end{subfigure}

    \caption{
    Effect of the number of pre-trained models and the number of low-quality models under the miniImageNet 5-way 1-shot setting.
    (a) Meta-testing accuracies with different numbers of pre-trained models.
    (b) Meta-testing accuracies with different numbers of low-quality pre-trained models.
    Bars indicate the standard error.
    }
    \label{fig:model_pool}
\end{figure*}

\subsection{Results}\label{result}

Tables~\ref{omniglot} and \ref{imagenet} presented the average accuracies (\%) of different methods under various task settings, along with the standard errors when using different unlabeled datasets; for methods that do not utilize unlabeled data, only the average accuracy is reported. Bold numbers indicated results that are not statistically different from the best-performing method on each dataset at the 5\% level according to a paired t-test. The proposed method achieved the highest accuracy. For most methods, accuracy increased reasonably as the support set size grew. The proposed method outperformed non-meta-learning baselines (Random, K-nearest neighbors, and Best pre-trained model). This result indicated the importance of sharing knowledge across tasks, and it is difficult to perform well with only a small number of training samples. The accuracy of the existing unsupervised meta-learning method (PsCo) was lower than that of the proposed method. This result implies that leveraging pre-trained models to construct diverse and target-relevant tasks is effective. The data-free meta-learning method (FREE) performed worse than the proposed method. Together with the training time results shown in Table~\ref{training_time}, this indicates that incorporating unlabeled data related to the target tasks not only improves performance but also avoids costly data recovery, thereby achieving substantial speed-ups. Although KD-MAML also leverages pre-trained models and unlabeled data for meta-learning, it ignores the differences in soft label quality and relies on hard-label updates in the inner loop, resulting in inferior performance compared to the proposed method. This further demonstrated that the performance improvements of the proposed method are not only due to the newly introduced setting but also attributable to the effectiveness of the method itself.

In practical applications, publicly available pre-trained models often originate from domains that are not perfectly aligned with the target domain, due to resource diversity and selection constraints. To reflect this realistic scenario, we conducted experiments under a more challenging multi-domain setting. In this setting, the pool of pre-trained models included not only those trained on the target domain (CUB or CIFAR-FS) but also models trained on a different domain (miniImageNet). This setting requires algorithms to extract useful knowledge while demonstrating robust generalization across multiple domains. As shown in Table~\ref{cross_domain}, the proposed method achieved the highest accuracy. The results showed that, even when pre-trained models are drawn from multiple domains, our method can effectively leverage them for robust meta-learning. By incorporating the proposed task-weighting mechanism, our method discerns the usefulness of tasks generated from different models, thereby enabling efficient meta-learning across domains. 

In real-world applications, the diversity of publicly available resources often leads to pre-trained models that differ in both architecture and scale. To evaluate this scenario, we further assessed our method in a cross-architecture setting, where each pre-trained model adopted a different architecture. Specifically, the architecture of each model was randomly selected from Conv4, Conv5 \citep{finn2017model, snell2017prototypical}, ResNet-10, and ResNet-18 \citep{he2016deep}. As shown in Table~\ref{cross_archi}, our method achieved the best performance among all comparative approaches and could be directly applied to cross-architecture scenarios without modification. These results demonstrated that our method effectively learns across heterogeneous model architectures. This robustness arises because our method makes no assumptions regarding the architecture or scale of the pre-trained models.

%We first evaluate the effect of the task-weighting mechanism. As shown in Table~\ref{ablation} ($\alpha = 0, \beta=1$),
We also conducted ablation studies to further investigate how each key component contributed to the performance of the proposed method.
We first evaluated the effectiveness of the task-weighting mechanism. As illustrated in Table~\ref{ablation} (w/o weight), removing this mechanism results in a decline in meta-testing performance, demonstrating its capability to accurately identify the usefulness of tasks generated by various pre-trained models, thereby enhancing performance on unseen tasks. We also examined the contributions of two factors within the task-weighting mechanism. Table~\ref{ablation} showed that removing either component ($\alpha = 0, \beta = 1$ or $\alpha = 1, \beta = 0$) leads to reduced meta-testing performance compared to the proposed method (Ours). This highlighted the significance of both components in effectively discerning the usefulness of various meta-training tasks. Next, we assessed the impact of modifying the inner optimization. Specifically, soft labels in the support set were converted into hard labels via argmax, and prototypes were computed as the mean of sample embeddings for each class. As shown in Table~\ref{ablation} (inner hard), using hard labels leads to a decrease in accuracy. This is because converting soft to hard labels may result in insufficient samples for some classes, and prototypes calculated by a few examples are typically less robust. Moreover, since hard labels contain less information than soft labels, this may also lead to a degradation in performance. Finally, we evaluated the effect of modifying the query loss (Eq.~\eqref{query_loss}) by replacing soft labels in the query set with hard labels and computing the loss using cross-entropy. As shown in Table~\ref{ablation} (outer hard), this change results in degraded performance. The reason is that soft labels contain richer information than hard labels, a well-established finding in the knowledge distillation \citep{gou2021knowledge, phuong2019towards, hinton2015distilling}.

%\begin{figure}[t]
  %\centering
    %\begin{subfigure}[b]{0.48\textwidth}
    %\centering
    %\includegraphics[width=\textwidth]{01_0.2.png}
    %\caption{$(\alpha,\beta) = (0, 1)$, Correlation coefficient: -0.25}
  %\end{subfigure}
  %\begin{subfigure}[b]{0.48\textwidth}
    %\centering
    %\includegraphics[width=\textwidth]{10_0.36.png}
    %\caption{$(\alpha,\beta) = (1, 0)$, Correlation coefficient: -0.40}
  %\end{subfigure}
  %\hfill
    %\begin{subfigure}[b]{0.48\textwidth}
    %\centering
    %\includegraphics[width=\textwidth]{0.50.5_0.54.png}
    %\caption{$(\alpha,\beta) = (0.5, 0.5)$, Correlation coefficient: -0.58}
    %\end{subfigure}
  %\caption{Relationship between task weights (computed using Eq.~\eqref{weight}) and cross-entropy with ground-truth labels for meta-training tasks constructed by Eq.~\eqref{y}. Each point represents a single task. Different subfigures correspond to different $\alpha$ and $\beta$ settings. Pearson correlation coefficients are also reported.}
  %\label{t-sne}
%\end{figure}

%\begin{figure}
    %\includegraphics[width=0.32\textwidth]{nm_1.png}
    %\caption{Meta-testing accuracies with varying numbers of pre-trained models under the miniImageNet 5-way 1-shot setting. The bar shows the standard error.}
    %\label{num_models}
%\end{figure}

\begin{figure*}[t]
  \centering
  \vspace{-2.5mm}
  \begin{subfigure}[t]{0.32\textwidth}
    \centering
    \includegraphics[width=\linewidth]{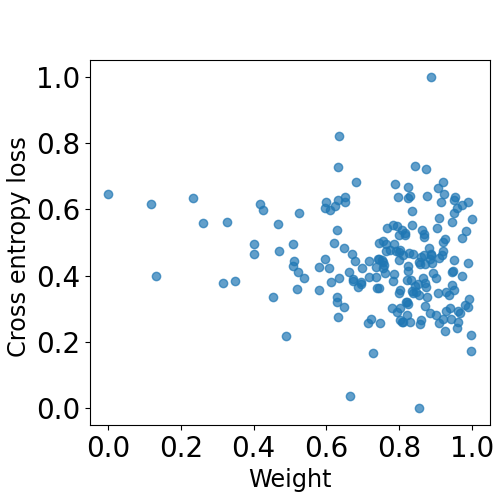}
    \caption{$(\alpha,\beta) = (0, 1)$, PCC: -0.15}
  \end{subfigure}
  \hfill
  \begin{subfigure}[t]{0.32\textwidth}
    \centering
    \includegraphics[width=\linewidth]{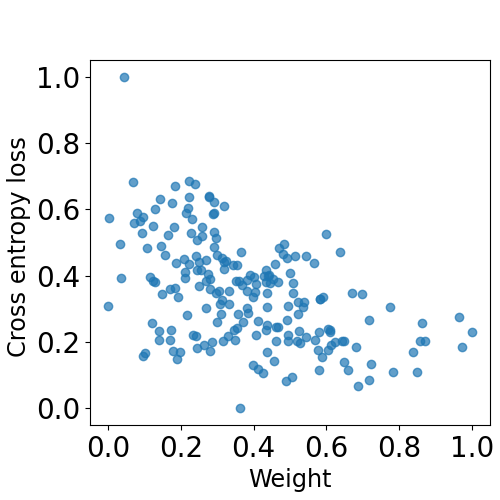}
    \caption{$(\alpha,\beta) = (1, 0)$, PCC: -0.33}
  \end{subfigure}
 \hfill
  \begin{subfigure}[t]{0.32\textwidth}
    \centering
    \includegraphics[width=\linewidth]{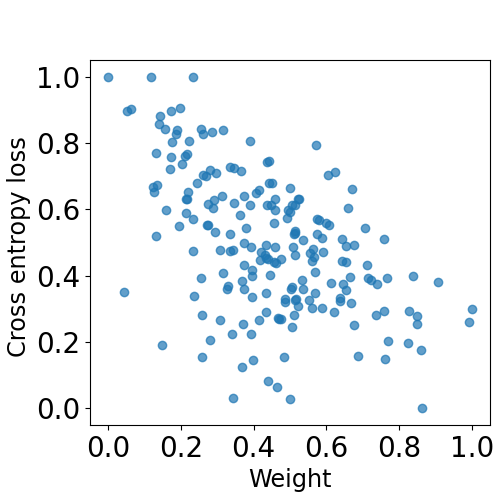}
    \caption{$(\alpha,\beta) = (0.5, 0.5)$, PCC: -0.56}
  \end{subfigure}
  \caption{
    (a--c) Relationship between task weights (computed using Eq.~\eqref{weight}) and cross-entropy with ground-truth labels for meta-training tasks constructed by Eq.~\eqref{y}. Each point represents a single task. Pearson correlation coefficients (PCC) are also reported.}
  \label{t-sne}
\end{figure*}

\begin{table}[t]
\centering
\caption{Average accuracy (\%) and standard error on miniImageNet under 5-way 1-shot and 5-way 5-shot settings. Ours-ProtoNet and Ours-MAML denote two instantiations of the proposed framework based on Prototypical Networks~\citep{snell2017prototypical} and MAML~\citep{finn2017model}, respectively.}\label{ours_maml}
\begin{tabular}{lcc}
\toprule
\textbf{Method} & {(5-way, 1-shot)} & {(5-way, 5-shot)} \\
\midrule
PsCo          & 31.7 $\pm$ 0.4   & 38.5 $\pm$ 1.8   \\
FREE          & 25.2             & 35.8             \\
KD-MAML       & 28.3 $\pm$ 1.7   & 36.3 $\pm$ 2.2   \\
Ours-ProtoNet & 33.6 $\pm$ 1.2   & 48.7 $\pm$ 0.9   \\
Ours-MAML     & 32.7 $\pm$ 1.6   & 48.1 $\pm$ 1.1   \\
\bottomrule
\end{tabular}
\end{table}

We conducted experiments under the 1-shot setting to analyze the sensitivity of the proposed task-weighting mechanism to the balance factors $\alpha$ and $\beta$, where $\beta = 1 - \alpha$. As shown in Figure~\ref{fig:alpha_beta}, using both entropy-based factors achieved better or comparable performance than using either factor alone, i.e., $\alpha = 0$ or $\alpha = 1$, in most cases across Omniglot, miniImageNet, and tieredImageNet. These results indicate that both factors in Eq.~\eqref{weight} are important for improving generalization to unseen tasks.

To further analyze the effectiveness of the weighting mechanism, we conducted experiments on miniImageNet under the 5-way setting. Specifically, we randomly sample 200 meta-training tasks from those constructed by Eq.\eqref{y}, compute the cross-entropy between the soft labels and ground-truth labels for each task, and calculate the corresponding task weight using Eq.~\eqref{weight}. A lower cross-entropy loss indicates that the soft labels are more consistent with the ground-truth labels. Note that ground-truth labels are not accessible during the meta-training phase. We then visualize the relationship between task weights and cross-entropy and compute their correlation coefficient. As shown in Figure~\ref{t-sne}(c), the task weight is negatively correlated with cross-entropy, with a correlation coefficient of -0.56. This suggests that tasks with higher weights tend to have soft labels more consistent with the ground-truth labels. These results validate the effectiveness of our weighting strategy: tasks with more accurate soft labels provide clearer supervision signals and more balanced class distributions, both of which are essential for the meta-training process. Additionally, we compare the same correlation under $\alpha = 1, \beta = 0$ and $\alpha = 0, \beta = 1$. Although correlation is observed in both cases, the absolute values of the correlation coefficients are substantially lower than those obtained with $\alpha = 0.5, \beta = 0.5$, further highlighting the necessity of incorporating both factors in Eq.~\eqref{weight}.

We also evaluated the effect of varying the number of pre-trained models, each independently trained on miniImageNet under the 5-way 1-shot setting. As shown in Figure~\ref{fig:num_models}, meta-testing accuracies improved progressively with the number of pre-trained models, up to approximately 700. This is because a larger pool of pre-trained models enables the construction of more diverse and target-relevant meta-training tasks. This helps the model acquire a more robust inductive bias, facilitating faster and more effective adaptation to unseen tasks. 

In practical applications, the diversity of publicly available resources often leads to the inclusion of low-quality models. To investigate the impact of low-quality models, we evaluated the effect of varying their proportion in the pre-trained model pool under the miniImageNet 5-way 1-shot setting. As described in Section~\ref{setup}, the miniImageNet pool already contained 3 low-quality models (with accuracies below 50\%). To further adjust the proportion of low-quality models, we replaced a subset of the pool with models initialized with random parameters. As Figure~\ref{fig:noisy_models} showed, meta-testing accuracy decreased as the proportion of low-quality models increased. Nevertheless, accuracy remained relatively stable at around 33–34\% when the number of low-quality models ranged between 3 and 30. Once the number exceeded 60, performance began to decline. As expected, accuracy dropped sharply when most of the pre-trained models were low-quality (close to 100), as reliable soft labels were difficult to obtain under such conditions. These results demonstrated that, even with a substantial proportion of low-quality models, the proposed task-weighting mechanism can effectively identify the usefulness of tasks generated from pre-trained models, thereby enabling efficient meta-learning.

We further evaluated whether the proposed framework is tied to a specific meta-learning algorithm. As shown in Table~\ref{ours_maml}, although Ours-ProtoNet achieved slightly higher average accuracies than Ours-MAML on miniImageNet under both the 5-way 1-shot and 5-way 5-shot settings, the differences were not statistically significant. In addition, Ours-MAML outperformed all baseline methods, indicating that the proposed framework can also improve a gradient-based meta-learning algorithm such as MAML. Since Ours-MAML and KD-MAML used the same meta-learning algorithm, the improvements of Ours-MAML over KD-MAML further demonstrated the effectiveness of the proposed soft-label task construction and task-weighting mechanism, rather than being attributable to ProtoNets alone.

\section{Conclusion}
We proposed a meta-learning setting that leverages both pre-trained models and unlabeled data. In this setting, our method constructs meta-training tasks by assigning soft labels from pre-trained models to unlabeled data, thereby avoiding the high computational cost and difficulties of data recovery. To further ensure effective meta-learning, we introduce a task-weighting mechanism based on confidence and class balance. We believe this work is an important step for learning from diverse information sources, and future research may extend this framework to broader tasks such as regression and reinforcement learning.

\section*{Funding statement}
This work was supported by JST SPRING, Grant Number JPMJSP2140, Japan.

\bibliographystyle{cas-model2-names}

\bibliography{cas-refs}

\end{document}